%% file: main.tex
\documentclass{article} 
\usepackage{iclr2015,times}
\usepackage{hyperref,url}
\usepackage{times,color,float}
\usepackage{latexsym}
\usepackage{amsmath,amsfonts,amssymb}
\usepackage{algorithm,algorithmic}
\usepackage{graphicx}
\usepackage{booktabs}
\usepackage{multirow}
\usepackage{tikz,tikz-qtree}
\usepackage{paralist}
\usepackage{subcaption}
\usepackage{mystuff-nlp}
\usepackage{enumitem}

\newcommand{\modelname}{\textsc{disco2}}

\newcommand{\uvec}[1]{\vec{u}_{#1}} 
\newcommand{\dvec}[1]{\vec{d}_{#1}} 
\newcommand{\fvec}[1]{\vec{\phi}_{#1}}
\newcommand{\uclass}[1]{\mat{A}_{#1}} 
\newcommand{\dclass}[1]{\mat{B}_{#1}} 
\newcommand{\fclass}[1]{\vec{\beta}_{#1}} 
\newcommand{\ucomp}{\mat{U}} 
\newcommand{\dcomp}{\mat{V}} 
\newcommand{\lchild}[1]{\ell(#1)} 
\newcommand{\rchild}[1]{r(#1)} 
\newcommand{\mytanh}[1]{\text{tanh}\left(#1\right)}
\newcommand{\example}[1]{\textit{#1}}
\renewcommand{\trans}[1]{\ensuremath{#1^{\top}}}

\usetikzlibrary{trees,positioning,arrows}

\newcommand{\vu}{\vec{u}}

\newcommand{\vum}{\vec{u}^{(m)}}
\newcommand{\vun}{\vec{u}^{(n)}}
\newcommand{\vd}{\vec{d}}

\newcommand{\vdm}{\vec{d}^{(m)}}
\newcommand{\vdn}{\vec{d}^{(n)}}
\tikzstyle{every node}=[font=\small]
\tikzstyle{up}=[edge from parent/.style={draw,latex-}]
\tikzstyle{down}=[edge from parent/.style={draw,-latex}]
\tikzstyle{nodraw}=[edge from parent/.style={}]

\title{Entity-Augmented Distributional Semantics for Discourse Relations}

\author{
Yangfeng Ji \& Jacob Eisenstein \\
School of Interactive Computing\\
Georgia Institute of Technology\\
Atlanta, GA 30332, USA \\
\texttt{\{jiyfeng,jacobe\}@gatech.edu} \\
}

\iclrfinalcopy 

\begin{document}

\maketitle

\begin{abstract}
Discourse relations bind smaller linguistic elements into coherent texts. However, automatically identifying discourse relations is difficult, because it requires understanding the semantics of the linked sentences. A more subtle challenge is that it is not enough to represent the meaning of each sentence of a discourse relation, because the relation may depend on links between lower-level elements, such as entity mentions. Our solution computes distributional meaning representations by composition up the syntactic parse tree. A key difference from previous work on compositional distributional semantics is that we also compute representations for entity mentions, using a novel downward compositional pass. Discourse relations are predicted not only from the distributional representations of the sentences, but also of their coreferent entity mentions. The resulting system obtains substantial improvements over the previous state-of-the-art in predicting implicit discourse relations in the Penn Discourse Treebank.
\end{abstract}

\input{intro}

\input{model}

\input{exp}


\bibliography{ref,cite-strings,cites,cite-definitions}
\bibliographystyle{iclr2015}
\end{document}

%% file: intro.tex

\section{Introduction}

The high-level organization of text can be characterized in terms of \textbf{discourse relations} between adjacent spans of text~\citep{knott1996data,mann1984discourse,webber1999discourse}. Identifying these relations has been shown to be relevant to tasks such as summarization~\citep{louis2010discourse,yoshida2014dependency}, sentiment analysis~\citep{somasundaran2009supervised}, and coherence evaluation~\citep{lin2011automatically}. While the Penn Discourse Treebank (PDTB) now provides a large dataset annotated for discourse relations~\citep{prasad2008penn}, the automatic identification of implicit discourse relations is a difficult task, with state-of-the-art performance at roughly 40\%~\citep{lin2009recognizing}.

One reason for this poor performance is that predicting implicit discourse relations is a fundamentally semantic task, and the relevant semantics may be difficult to recover from surface level features. For example, consider the discourse relation between the following two sentences in Example (1), where a discourse connector like ``\emph{because}'' seems appropriate to indicate the relationship. However, without discourse connector, there is little surface information to signal the relationship. We address this issue by applying a discriminatively-trained model of compositional distributional semantics to discourse relation classification~\citep{socher2013recursive,baroni2013frege}. The meaning of each sentence is represented as a vector~\citep{turney2010frequency}, which is computed through a series of compositional operations over the parse tree. 

\begin{tabular}{llll}
  Example (1) : & Bob gave Tina the burger. & Example (2) : & Bob gave Tina the burger.\\
  & She was hungry. & & He was hungry.\\
\end{tabular}

We further argue that purely vector-based representations on sentences are insufficiently expressive to capture discourse relations. To see why, consider what happens in Example (2), where a tiny change is made based on Example (1). After changing the subject of the second sentence to Bob, the original discourse relation seems no longer holding in Example (2). But despite the radical difference in meaning, the distributional representation of the second sentence will be almost unchanged: the syntactic structure remains identical, and the words ``\example{he}'' and ``\example{she}'' have very similar word representations. We address this issue by computing vector representations not only for each sentence, but also for each coreferent entity mention within the sentences. These representations are meant to capture the \emph{role} played by the entity in the text. We compute entity-role representations using a novel feed-forward compositional model, which combines \emph{upward} and \emph{downward} passes through the syntactic structure. Representations for these coreferent mentions are then combined into a classification model, and help to predict the implicit discourse relation. In combination, our approach achieves a 3\% improvement in accuracy over the best previous work~\citep{lin2009recognizing} on the second-level discourse relation identification in the PDTB.\footnote{For more details, please refer to the long version of this paper \citep{ji2015one}}.

Our model requires a syntactic parse tree, which is produced automatically from the Stanford CoreNLP parser~\cite{klein2003accurate}. A reviewer asked whether it might be better to employ a left-to-right recurrent neural network, which would obviate the need for this language-specific resource. While it would clearly be preferable to avoid the use of language-specific resources whenever possible, we think this approach is unlikely to succeed in this case. A key difference between language and other types of data is that language has inherent recursive structure. A rich literature in both linguistics and natural language processing elaborates on the close relationship between (recursively-structured) syntax and semantics. Therefore, we see strong theoretical evidence --- as well as practical evidence from the history of natural language processing --- that syntactic parse structures are central to capturing the meaning in text. 

Regarding the multilingual question, there are now accurate parsers and annotated treebanks for dozens of languages,\footnote{http://universaldependencies.github.io/docs/\#language-other}
and training accurate parsers for ``low resource'' languages is a hot research topic, with substantial interest from both industry and academia. Languages differ substantially in the importance of word ordering, with English emphasizing word order more than most other languages~\citep{bender2013linguistic}. To our knowledge, it is an open question as to whether left-to-right recurrent neural networks will successfully extract meaning in languages where word order is more free.

%% file: model.tex

\section{Entity augmented distributional semantics for Relation Identification}\label{sec:model}

We briefly describe our approach to entity-augmented distributional semantics and to discourse relation identification. Our relation identification model is named as~\modelname, since it is a {\bf dis}tributional {\bf com}positional approach to {\bf disco}urse relations.

\subsection{Entity augmented distributional semantics}
The entity-augmented distributional semantics includes two passes in composition procedure: the upward pass for distributional representation of sentence, while the downward pass for distributional representation of entities shared between sentences.
\paragraph{Upward pass}
Distributional representations for sentences are computed in a feed-forward \emph{upward} pass: each non-terminal in the binarized syntactic parse tree has a $K$-dimensional distributional representation that is computed from the distributional representations of its children, bottoming out in representations of individual words. We follow the Recursive Neural Network (RNN) model proposed by \citet{socher2011parsing}. Specifically, for a given parent node $i$, we denote the left child as $\lchild{i}$, and the right child as $\rchild{i}$. We compose their representations to obtain, 
$\vu_i = \mytanh{\ucomp [\uvec{\lchild{i}}; \uvec{\rchild{i}}]}$, where $\mytanh{\cdot}$ is the element-wise hyperbolic tangent function, and $\ucomp\in\mathbb{R}^{K\times 2K}$ is the upward composition matrix. We apply this compositional procedure from the bottom up, ultimately obtaining the sentence-level representation $\vu_0$. 

\paragraph{Downward pass} As seen in the contrast between Examples (1) and (2), a model that uses a single vector representation for each sentence would find little to distinguish between ``\example{she was hungry}'' and ``\example{he was hungry}''. It would therefore almost certainly fail to identify the correct discourse relation for at least one of these cases, which requires tracking the roles played by the entities that are coreferent in each pair of sentences. To address this issue, we augment the representation of each sentence with additional vectors, representing the semantics of the role played by each coreferent entity in each sentence. Rather than represent this information in a logical form --- which would require robust parsing to a logical representation --- we represent it through additional distributional vectors. The role of a constituent $i$ can be viewed as a combination of information from two neighboring nodes in the parse tree: its parent $\rho(i)$, and its sibling $s(i)$. We can make a downward pass, computing the downward vector $\vd_i$ from the downward vector of the parent $\vd_{\rho(i)}$, and the \textbf{upward} vector of the sibling $\vu_{s(i)}$: $\vd_i = \mytanh{\dcomp [\dvec{\rho(i)}; \uvec{s(i)}]}$, where $\dcomp\in\mathbb{R}^{K\times 2K}$ is the downward composition matrix. The base case of this recursive procedure occurs at the root of the parse tree, which is set equal to the upward representation, $\vd_0 \triangleq \vu_0$. 

\subsection{Relation identification model}
To predict the discourse relation between an sentence pair $(m,n)$, the decision function is a sum of bilinear products,
\begin{equation}
  \label{eq:decision}
  {\small
    \begin{split}
    \psi(y) & = \trans{(\vum_0)} \uclass{y} \vun_{0} + \sum_{i,j \in \set{A}(m,n)} 
    \trans{(\vdm_i)}\dclass{y} \vdn_j + \trans{\fclass{y}}\fvec{(m,n)} + b_y,\\
    \end{split}
  }
\end{equation}
where the predicted relation is given by $\hat{y}=\argmax{y \in \set{Y}} \psi(y)$, and $\uclass{y},\dclass{y} \in \mathbb{R}^{K\times K}$ are the classification parameters for relation $y$. A scalar $b_{y}$ is used as the bias term for relation $y$, and $\set{A}(m,n)$ is the set of coreferent entity mentions shared among the sentence pair $(m,n)$. For the cases where there are no coreferent entity mentions between two sentences, $\set{A}(m,n) = \varnothing$, the classification model considers only the upward vectors at the root. We also use the \emph{surface features} vector $\fvec{(m,n)}$ in the decision function, as we find that, this approach outperforms prior work on the classification of implicit discourse relations in the PDTB, when combined with a small number of surface features.

%% file: exp.tex

\section{Experiments}
\input{tab-pdtb}

We evaluate our approach on the implicit discourse relation identification in the Penn Discourse Treebank (PDTB). PDTB relations may be \emph{explicit}, meaning that they are signaled by discourse connectives (e.g., \example{because}); alternatively, they may be \emph{implicit}, meaning that the connective is absent. We focus on the more challenging problem of classifying implicit discourse relations. Aiming to build a discourse parser in future, we follow the same experimental setting proposed by \citet{lin2009recognizing}, and evaluate our relation identification model on the \emph{second-level} relation types.

We run the Stanford parser \citep{klein2003accurate} and the Berkeley coreference system \citep{DurrettKlein2013} to obtain syntactic trees and coreference results respectively. In the PDTB, each discourse relation is annotated between two argument spans. For non-sentence argument span, we identify the syntactic subtrees with the span, and construct a right-branching superstructure to unify them into a tree. 

Table~\ref{tab:results-pdtb} presents results for multiclass identification of second-level PDTB relations. As shown in lines 5 and 6, \modelname~outperforms the prior state-of-the-art (line 1). The strongest performance is obtained by including the entity distributional semantics, with a 3.4\% improvement over the accuracy reported by \citet{lin2009recognizing} ($p < .05$). The improvement over our reimplementation of this work (line 2) is even greater, which shows how the distributional representation provides additional value over the surface features. The contribution of entity semantics is shown in Table~\ref{tab:results-pdtb} by the accuracy differences between lines 3 and 4, and between lines 5 and 6. 

\section{Conclusion}
Discourse relations are determined by the meaning of their arguments, and progress on discourse parsing therefore requires computing representations of the argument semantics. We present a compositional method for inducing distributed representations not only of discourse arguments, but also of the entities that thread through the discourse.
By jointly learning the relation classification weights and the compositional operators, this approach outperforms prior work based on hand-engineered surface features.  More discussion and experimental results can be found in a forthcoming journal paper~\citep{ji2015one}.

%% file: tab-pdtb.tex

\begin{table*}
  \centering
  {\small
  \begin{tabular}{lllll}
    \toprule
    Model &  +Entity semantics & +Surface features & $K$ & Accuracy(\%) \\
    \midrule
    \emph{Prior work}\\
    1. \cite{lin2009recognizing}  &   & Yes  &  & 40.2\\[0.5em]
    \emph{Our work}\\
    2. Surface feature model  &   & Yes  &  & 39.69\\[0.5em]
    3. \modelname   & No & No & 50 & 36.98 \\
    4. \modelname   & Yes & No & 50 & 37.63\\[0.3em]
    5. \modelname   & No & Yes  & 50 & 42.53\\
    6. \modelname   & Yes & Yes  & 50 & {\bf 43.56}$^\ast$\\
    \bottomrule
    \multicolumn{4}{l}{$^\ast$ signficantly better than \citet{lin2009recognizing} with $p<0.05$}\\
  \end{tabular}
  }
  \caption{Experimental results on multiclass classification of second-level discourse relations. The results of \citet{lin2009recognizing} are shown in line 1; the results for our reimplementation of this system are shown in line 2.}
  \label{tab:results-pdtb}
\end{table*}